\documentclass[10pt,twocolumn,letterpaper]{article}

\usepackage[pagenumbers]{cvpr} 

\usepackage{graphicx}
\usepackage{amsmath}
\usepackage{amssymb}
\usepackage{booktabs}
\usepackage{multirow}
\usepackage[table]{xcolor}

\usepackage[pagebackref,breaklinks,colorlinks]{hyperref}

\usepackage[capitalize]{cleveref}
\crefname{section}{Sec.}{Secs.}
\Crefname{section}{Section}{Sections}
\Crefname{table}{Table}{Tables}
\crefname{table}{Tab.}{Tabs.}

\def\confName{arxiv}
\def\confYear{2025}

\begin{document}

\title{ZonUI-3B: Competitive GUI Grounding with a 3B VLM Trained on a Single Consumer GPU}

\author{
ZongHan Hsieh \quad ShengJing Yang \quad Tzer-Jen Wei\\
DeepCAT Lab, National Yang Ming Chiao Tung University\\
{\tt\small \{zonghan.ai12, billy004104.ai12, tjwei\}@nycu.edu.tw}
}
\maketitle

\begin{abstract}
In this paper, we present ZonUI-3B, a lightweight Vision-Language Model (VLM) that can be fully trained on a single consumer-grade GPU (RTX 4090) while delivering performance comparable to significantly larger models on GUI grounding tasks. The model incorporates several key innovations: (i) combine cross-platform, multi-resolution dataset of 24K examples from diverse sources including mobile, desktop, and web GUI screenshots to effectively address data scarcity in high-resolution desktop environments; (ii) a two-stage fine-tuning strategy, where initial cross-platform training establishes robust GUI understanding, followed by specialized fine-tuning on high-resolution data to significantly enhance model adaptability; and (iii) data curation and redundancy reduction strategies, demonstrating that randomly sampling a smaller subset with reduced redundancy achieves performance comparable to larger datasets, emphasizing data diversity over sheer volume. Empirical evaluation on standard GUI grounding benchmarks—including ScreenSpot, ScreenSpot-v2, and the challenging ScreenSpot-Pro—highlights ZonUI-3B's exceptional accuracy, achieving 84.9\% on ScreenSpot and 86.4\% on ScreenSpot-v2, surpassing prior models under 4B parameters. Ablation studies validate the critical role of balanced sampling and two-stage fine-tuning in enhancing robustness, particularly in high-resolution desktop scenarios. The ZonUI-3B model and related resources are available at: \url{https://github.com/Han1018/ZonUI-3B}.
\end{abstract}

\section{Introduction}
\label{sec:intro}

\begin{figure}[t]
  \centering
  \includegraphics[width=\linewidth]{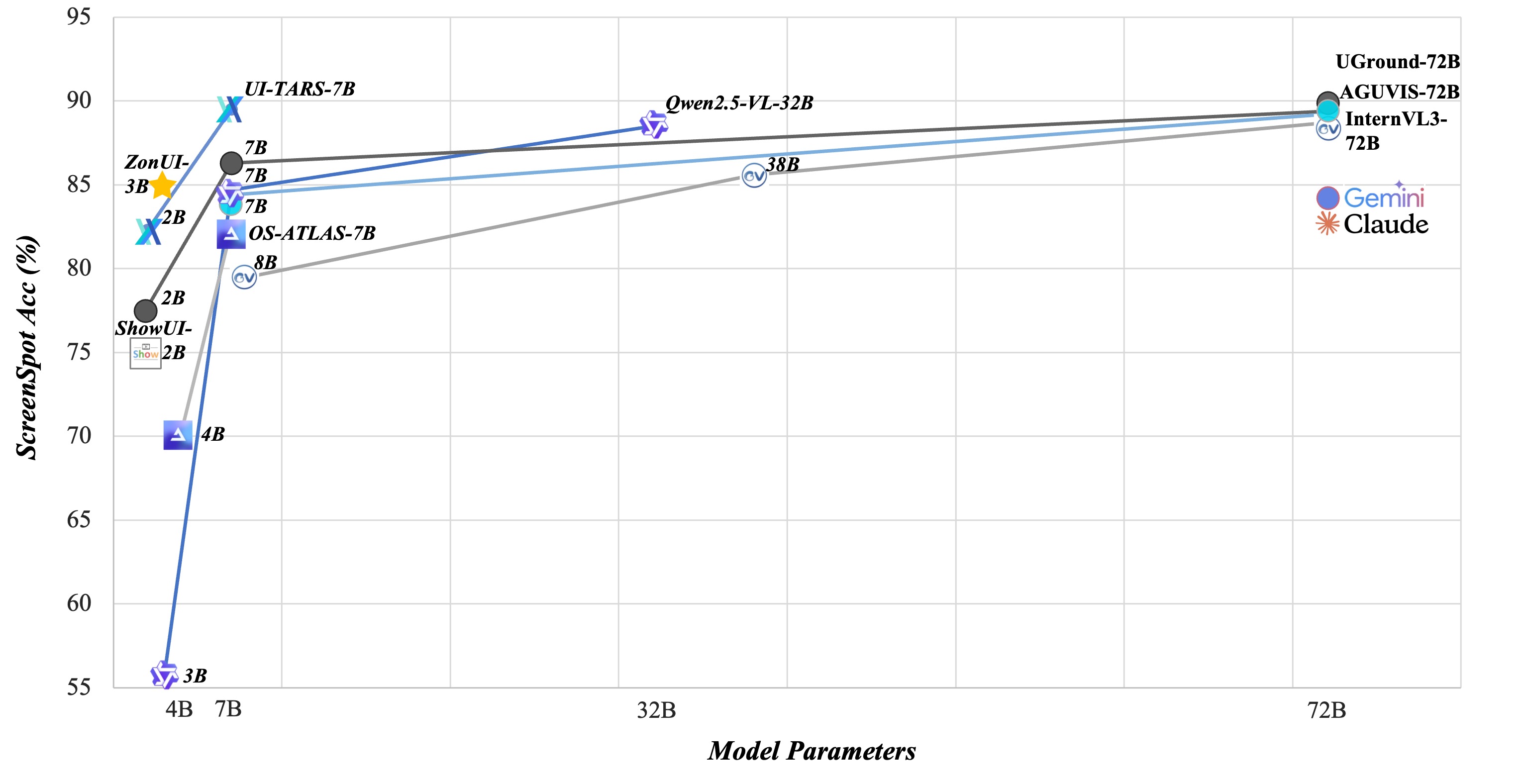}
  \caption{Overall ScreenSpot benchmark accuracy across model sizes.}
  \label{fig:ss-performance}
\end{figure}

Graphical User Interface (GUI) grounding—the task of locating the correct UI element on a screen given a natural language instruction—is a foundational capability for building intelligent GUI agents. While large-scale Vision-Language Models (VLMs) with over 7B parameters have achieved strong performance on this task \cite{xu2024aguvis, gou2024uground, wu2024osatlas}, their training requirements are prohibitively expensive, making them inaccessible for many researchers and practitioners without high-end hardware. This has sparked growing interest in developing compact models that retain high grounding accuracy \cite{lin2024showui, qin2025uitars}.

Early work in this direction, such as ShowUI \cite{lin2024showui}, demonstrated that lightweight models can achieve competitive results in zero-shot settings using only 2B parameters. However, achieving consistent performance across diverse GUI environments—including high-resolution desktop and web interfaces—remains a significant challenge. Smaller models often struggle with generalization when faced with dense layouts, varied screen resolutions, and UI patterns that deviate from training distributions. These limitations are exacerbated by three factors: (1) limited resolution diversity in existing datasets, which restricts robustness to large-scale interfaces \cite{gou2024uground}; (2) high variability in GUI structure and element styles, which introduces sensitivity to layout changes and visual distortions \cite{cheng2024seeclick}; and (3) data imbalance across platforms, particularly the under-representation of high-resolution desktop examples compared to mobile data \cite{chai2025amex}.

\begin{figure}[t]
  \centering
  \includegraphics[width=\linewidth]{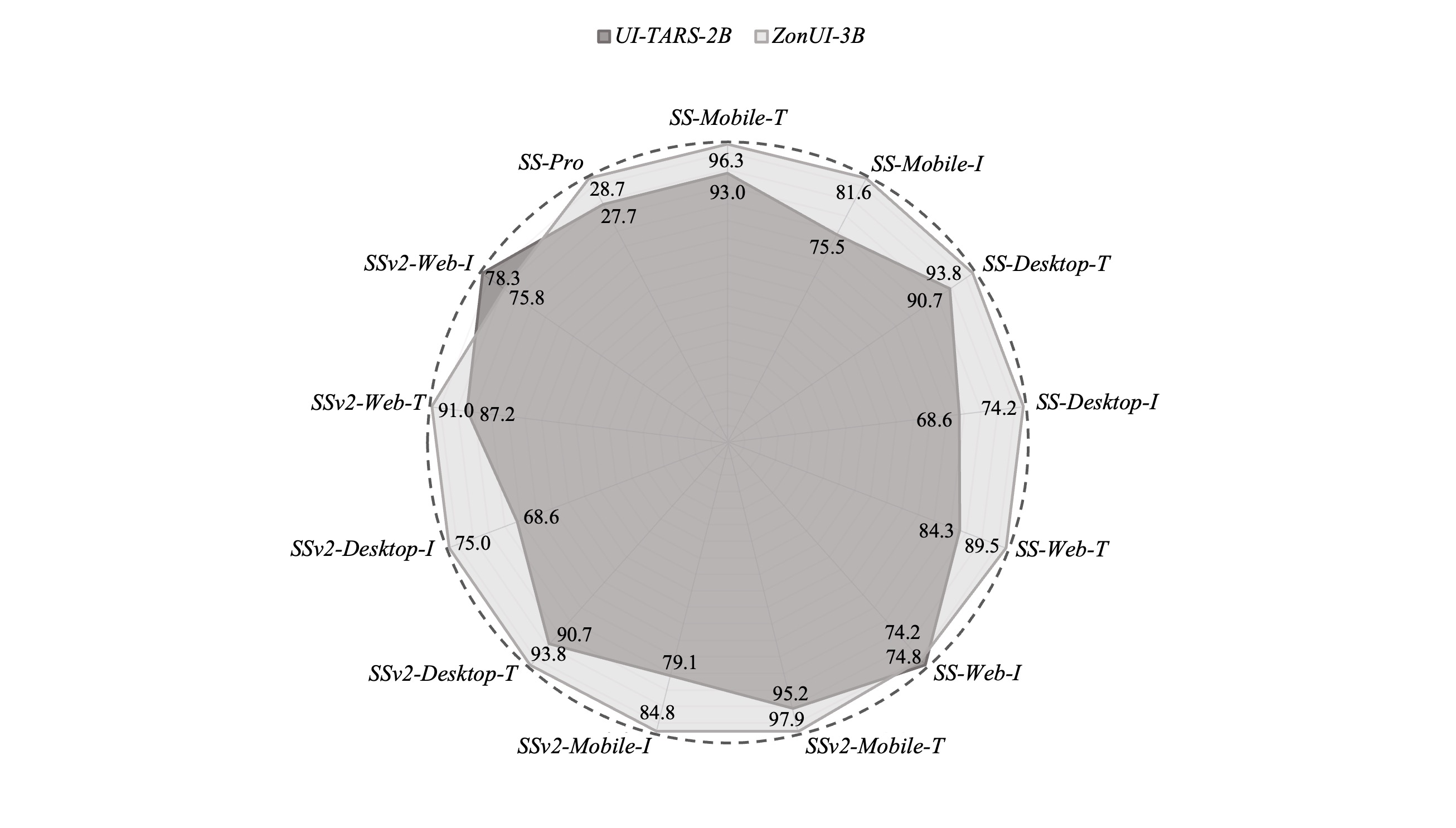}
  \caption{Per-category accuracy on ScreenSpot, ScreenSpot-v2, and ScreenSpot-Pro.}
  \label{fig:ss-compare}
\end{figure}

To address these issues, we propose ZonUI-3B, a lightweight 3B-parameter vision-language model optimized for GUI grounding in cross-platform and resolution-diverse environments. Despite its compact size, the model achieves accuracy on par with 7B-scale baselines on standard benchmarks \cite{gou2024uground, wu2024osatlas, xu2024aguvis}, demonstrating that model performance can be significantly improved through targeted training strategies and data selection rather than scale alone (see Figures~\ref{fig:ss-performance} and \ref{fig:ss-compare}).

Our contributions are three-fold:

(1) We construct a cross-platform, multi-resolution training corpus that combines open-source data from mobile, desktop, and web GUI environments \cite{lin2024showui, gou2024uground, chai2025amex}. This design improves coverage across screen types and mitigates platform-specific data scarcity.

(2) We introduce a two-stage fine-tuning strategy comprising platform-general pretraining followed by resolution-focused specialization. This staged approach improves grounding robustness across heterogeneous layouts without introducing architectural changes.

(3) We show that training on a compact 24K-example subset—randomly sampled from a redundant corpus—achieves comparable accuracy to full-dataset training. This highlights the importance of data diversity and sampling efficiency in lightweight model training.
\section{Data Recipe}
\subsection{Cross-Platform, Multi-Resolution Dataset Integration}

To support robust GUI grounding across diverse platforms and interface resolutions, we construct a unified training dataset by consolidating multiple publicly available GUI grounding sources. This integrated corpus is curated to maximize both platform diversity and resolution variation, which are critical for enabling generalization across real-world screen environments.

The dataset merges samples from ShowUI-Web and UGround-WebHybrid to capture web interfaces \cite{lin2024showui, gou2024uground}, AMEX to represent mobile applications \cite{chai2025amex}, and ShowUI-Desktop to cover desktop layouts \cite{lin2024showui}. The UGround component contributes multi-resolution web screenshots, synthesized at various resolutions and aspect ratios, including compact mobile views (e.g., 448×448) and full-scale desktop formats (e.g., 1344×1344). Some of these screenshots simulate mobile-style web layouts to enhance cross-device generalization \cite{gou2024uground}.

This combined dataset achieves dual diversity: broad platform coverage (Android, iOS, web, and desktop) and multi-resolution across GUI screenshots. In contrast to previous datasets—where ShowUI emphasized cross-platform variety but lacked resolution depth \cite{lin2024showui}, and UGround provided resolution richness but was web-centric \cite{gou2024uground}—our integrated corpus strikes a balance between breadth and fidelity. Despite its compact size, the resulting dataset is highly representative and information-dense, offering a practical foundation for efficient training of lightweight models.

\subsection{Data Diversification and Redundancy Reduction}

In constructing the training dataset, a key observation is that expanding dataset size beyond a certain point yields limited gains due to redundancy across GUI screenshots. Interfaces sourced from the same application or website often exhibit minor visual differences—such as static text updates or slight layout shifts—that contribute minimal new learning signal. This redundancy can reduce training efficiency and unnecessarily inflate resource consumption \cite{lin2024showui, chai2025amex}.

To address this, a random sampling strategy was applied to reduce the dataset volume without manual filtering or semantic clustering. Empirical results show that training with approximately one-seventh of the full dataset (16.1K examples) achieves comparable grounding accuracy to using the complete 120K sample set. For instance, performance on the ScreenSpot benchmark reached 82.8\% with the reduced dataset, versus 82.9\% with the full set—indicating that roughly 100K samples offered marginal utility.

\begin{figure*}[t]
    \centering
    \includegraphics[width=0.95\textwidth]{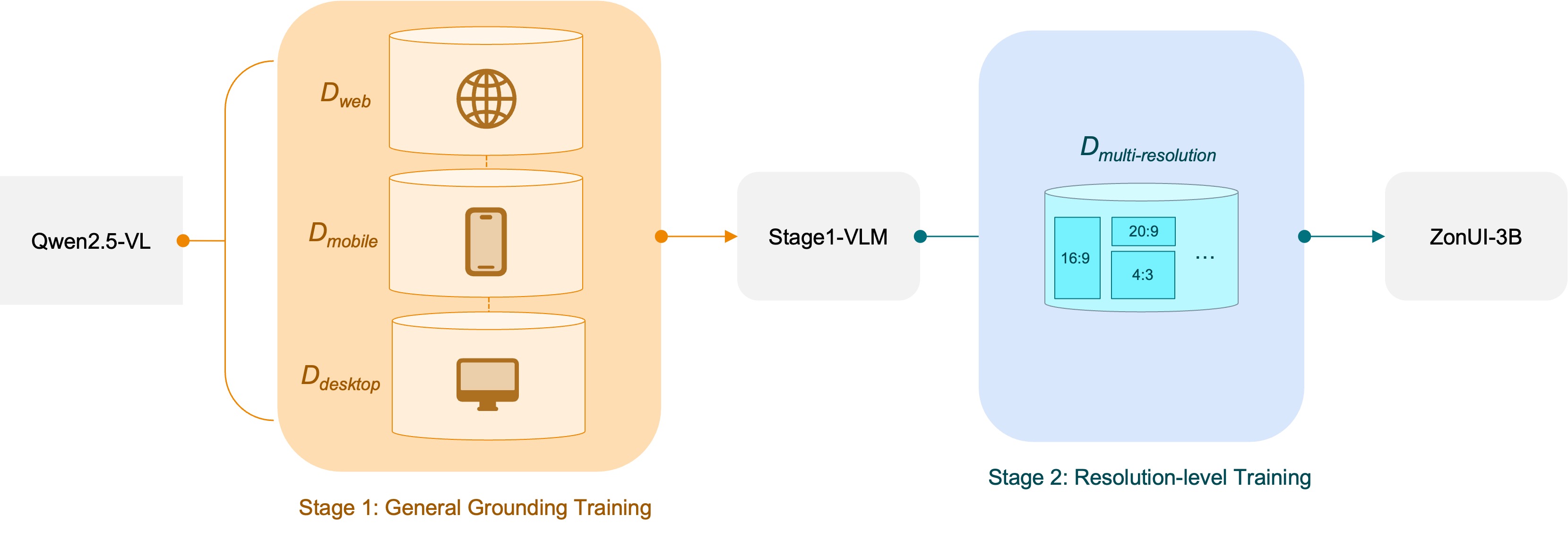}
    \caption{Overview of two-stage fine-tuning pipeline. Stage~1 builds platform-general GUI grounding ability across mobile, web, and desktop. Stage~2 focuses on resolution diversity, adapting the model on high-resolution data.}
    \label{fig:training-pipeline}
\end{figure*}

Analysis of individual sources reveals that redundancy stems from repeated GUI patterns: ShowUI-Web includes multiple captures of single webpage differing only in transient content \cite{lin2024showui}; AMEX mobile samples frequently reuse shared UI templates across task variants \cite{chai2025amex}. The random sampling process, while simple, effectively reduces duplication and yields a training set with higher information density.

By prioritizing minimizing redundancy, the resulting dataset supports efficient training of lightweight models on a single GPU, and establishes a practical baseline for scaling under data-limited or compute-constrained scenarios.

\section{Methodology}
\subsection{Lightweight Model Architecture}

The proposed model builds upon the Qwen2.5-VL-3B architecture \cite{bai2025qwen25vl}, a publicly available 3B-parameter vision-language model designed for efficient multi-modal understanding. This architecture supports dynamic input resolutions and multi-scale visual features, and includes native support for absolute coordinate grounding, making it a suitable backbone for GUI element localization. Its compact parameter size enables full fine-tuning on mainstream hardware (e.g., a single RTX 4090), offering a practical foundation for deployment in resource-limited environments.

To enable task-specific adaptation with minimal computational overhead, we adopt Low-Rank Adaptation (LoRA) \cite{hu2021lora}, which introduces trainable low-rank matrices into frozen transformer layers. This approach substantially reduces memory usage and training cost without modifying the base model’s architecture. Inference remains efficient, as LoRA parameters can be merged post-training, incurring no additional runtime latency or resource demands.

Throughout training, the architecture remains unaltered. The model processes raw GUI screenshots and natural language instructions without reliance on auxiliary inputs such as HTML structures, accessibility trees, or OCR modules. By preserving architectural simplicity, the approach facilitates controlled analysis of training strategies and dataset composition while maintaining compatibility with real-world deployment constraints.

\begin{table*}[ht]
\centering
\caption{Comparison of various models on ScreenSpot and ScreenSpot-v2}
\label{tab:screenspot}
\resizebox{0.7\textwidth}{!}{
\setlength{\tabcolsep}{5pt}
\renewcommand{\arraystretch}{0.9}
\begin{tabular}{lcccccccc}
\toprule
\multirow{2}{*}{\textbf{Model}} &
\multicolumn{4}{c}{\textbf{ScreenSpot Accuracy (\%)}} &
\multicolumn{4}{c}{\textbf{ScreenSpot-v2 Accuracy (\%)}} \\
\cmidrule(lr){2-5} \cmidrule(lr){6-9}
 & Mobile & Desktop & Web & \textbf{Avg.} & Mobile & Desktop & Web & \textbf{Avg.} \\
\midrule
\rowcolor{gray!15}\multicolumn{9}{l}{\textit{General Models}} \\
Qwen2.5-VL-3B~\cite{bai2025qwen25vl}    & -- & -- & -- & 55.5 & -- & -- & -- & -- \\
Qwen2.5-VL-7B~\cite{bai2025qwen25vl}    & -- & -- & -- & 84.7 & -- & -- & -- & -- \\
InternVL3-8B~\cite{zhu2025internvl3}    & -- & -- & -- & 79.5 & -- & -- & -- & 81.4 \\
Gemini-2 Flash~\cite{bai2025qwen25vl}   & -- & -- & -- & 83.0 & -- & -- & -- & -- \\
Claude3.5 Sonnet~\cite{bai2025qwen25vl} & -- & -- & -- & 84.0 & -- & -- & -- & -- \\
\midrule
\rowcolor{gray!15}\multicolumn{9}{l}{\textit{GUI-specific Models}} \\
CogAgent-18B~\cite{hong2023cogagent}     & 45.5 & 47.1 & 49.5 & 47.4 & -- & -- & -- & -- \\
SeeClick-9.6B~\cite{cheng2024seeclick}   & 65.0 & 51.0 & 44.1 & 53.4 & 64.5 & 49.7 & 43.8 & 55.1 \\
ShowUI-2B~\cite{lin2024showui}           & 80.3 & 70.4 & 74.2 & 74.9 & 83.7 & 69.1 & 72.6 & 77.3 \\
UGround-V1-2B~\cite{gou2024uground}      & 80.7 & 77.2 & 75.1 & 77.7 & -- & -- & -- & -- \\
UI-TARS-2B~\cite{qin2025uitars}          & 85.0 & 81.4 & 79.8 & 82.3 & 87.1 & 79.6 & 82.7 & \underline{84.7} \\
OSAtlas-7B~\cite{wu2024osatlas}          & 82.9 & 77.2 & \textbf{82.5} & 82.4 & 85.4 & 77.1 & \textbf{83.9} & 84.1 \\
Aguvis-7B~\cite{xu2024aguvis}            & \underline{86.6} & 80.4 & 81.7 & 84.4 & -- & -- & -- & -- \\
\midrule
\textbf{ZonUI-3B(Ours)} & \textbf{88.9} & \textbf{84.0} & \underline{81.8} & \textbf{84.9} & \textbf{91.3} & \textbf{84.4} & \underline{83.4} & \textbf{86.4} \\
\bottomrule
\end{tabular}
}
\end{table*}

\subsection{Two-Stage Fine-Tuning Strategy}

To enhance the model’s generalization across platforms and resolutions, we adopt a two-stage fine-tuning strategy that separates the learning of platform-invariant GUI semantics from resolution-specific adaptation. This staged design allows the model to first acquire broad grounding capabilities before specializing to high-resolution interface characteristics.

\textbf{Stage 1: Cross-Platform Pretraining.}
The model is first fine-tuned on a mixture of GUI data from mobile, web, and desktop sources \cite{lin2024showui, chai2025amex}. To mitigate the natural imbalance across platforms—particularly the underrepresentation of desktop samples—we apply a class-balanced sampling scheme that increases the frequency of desktop data during training. This ensures uniform exposure across device types and prevents bias toward mobile or web UIs. The objective of this stage is to establish a general understanding of common GUI components (e.g., text buttons, icons, menus) and natural language instructions across diverse environments.

\textbf{Stage 2: Resolution-Focused Specialization.}
The second stage further adapts the model using a subset of high- and multi-resolution screenshots, primarily drawn from UGround’s web-hybrid dataset \cite{gou2024uground}. This dataset introduces resolution and aspect ratio diversity by randomly selecting image sizes and simulating various device-specific web renderings, including mobile-style page layouts. Such variation enables the model to learn interface patterns across different screen scales and viewports—ranging from full-page desktop layouts to compact mobile-formatted webpages. The goal is to improve robustness under dense visual conditions, small clickable targets, and interface variability. Since Stage 1 has already built a stable grounding foundation, this specialization stage focuses solely on resolution-specific tuning without degrading cross-platform generality.

Throughout both stages, the model architecture remains unchanged. The same Qwen2.5-VL-3B backbone is retained~\cite{bai2025qwen25vl}, and no auxiliary components or data modalities are introduced. This two-stage schedule yields consistent gains in grounding accuracy, particularly for desktop and web platforms. It provides a lightweight yet effective mechanism to enhance model adaptability under diverse screen conditions without increasing model complexity or training cost (see Figure~\ref{fig:training-pipeline}).

\subsection{Efficient Training Pipeline and Comparison to Prior Work}

We present a streamlined training pipeline that emphasizes simplicity, data efficiency, and computational accessibility. In contrast to recent GUI agent frameworks that introduce complex architectures or rely on large-scale data acquisition and reinforcement learning infrastructure, our method demonstrates that a standard supervised fine-tuning setup—when paired with appropriate data and training strategy—is sufficient to achieve strong grounding performance.

Several recent models highlight this contrast. UI-TARS \cite{qin2025uitars} incorporates system-level reasoning modules, long-horizon memory, and iterative RL-based data collection across distributed environments. While these components are effective, they significantly increase training complexity and resource requirements. Similarly, UGround \cite{gou2024uground} relies on a massive-scale dataset (over 1.3 million screenshots and 10 million annotated elements) to achieve high coverage, requiring extensive data curation and storage capacity.

Our method, in comparison, relies on two orders of magnitude fewer training samples. The training data is carefully selected to maximize interface diversity and resolution variation, enabling strong generalization without exhaustive coverage. This compact design not only improves training efficiency but also reduces the infrastructure barrier for reproducing or extending the model.

Compared to ShowUI \cite{lin2024showui}, which also targets lightweight GUI grounding using a 2B model and 256K training examples, ZonUI-3B further simplifies the pipeline by avoiding custom architectural changes. Whereas ShowUI introduced a UI-guided token pruning module to reduce visual redundancy, our model benefits implicitly from the resolution diversity of the dataset, allowing the base architecture to focus attention on relevant interface regions without external guidance.

Our training process is kept minimal: fine-tuning a 3B backbone using LoRA-based adaptation \cite{hu2021lora} and gradient accumulation on a single RTX 4090 GPU. No auxiliary modules, external tools, or architectural modifications are introduced.

This methodology demonstrates that effective grounding performance can be achieved through careful data construction and staged adaptation, without scaling model size or pipeline complexity. The resulting model reaches performance levels comparable to 7B baselines \cite{wu2024osatlas, xu2024aguvis}, illustrating the practical potential of lightweight, well-optimized GUI grounding systems.

\begin{table*}[ht]
\centering
\caption{Comparison of various models on ScreenSpot-Pro. Most results are collected from official papers and leaderboard~\cite{li2025screenspotpro}}
\label{tab:screenspot_pro}
\setlength{\tabcolsep}{3pt}
\renewcommand{\arraystretch}{1.0}
\resizebox{\textwidth}{!}{%
\begin{tabular}{l*{21}{c}}
\toprule
\multirow{2}{*}{\textbf{Agent Model}} &
\multicolumn{3}{c}{\textbf{Development}} &
\multicolumn{3}{c}{\textbf{Creative}} &
\multicolumn{3}{c}{\textbf{CAD}} &
\multicolumn{3}{c}{\textbf{Scientific}} &
\multicolumn{3}{c}{\textbf{Office}} &
\multicolumn{3}{c}{\textbf{OS}} &
\multicolumn{3}{c}{\textbf{Avg.}} \\
\cmidrule(lr){2-4}\cmidrule(lr){5-7}\cmidrule(lr){8-10}
\cmidrule(lr){11-13}\cmidrule(lr){14-16}\cmidrule(lr){17-19}
\cmidrule(lr){20-22}
& Text & Icon & Avg. & Text & Icon & Avg. & Text & Icon & Avg.
& Text & Icon & Avg. & Text & Icon & Avg. & Text & Icon & Avg.
& Text & Icon & Avg.\\
\midrule
QwenVL-7B~\cite{li2025screenspotpro}          & 0.0 & 0.0 & 0.0 & 0.0 & 0.0 & 0.0 & 0.0 & 0.0 & 0.0 & 0.7 & 0.0 & 0.4 & 0.0 & 0.0 & 0.0 & 0.0 & 0.0 & 0.0 & 0.1 & 0.0 & 0.1 \\
GPT-4o~\cite{li2025screenspotpro}                & 1.3 & 0.0 & 0.7 & 1.0 & 0.0 & 0.6 & 2.0 & 0.0 & 1.5 & 2.1 & 0.0 & 1.2 & 1.1 & 0.0 & 0.9 & 0.0 & 0.0 & 0.0 & 1.3 & 0.0 & 0.8 \\
SeeClick              & 0.6 & 0.0 & 0.3 & 1.0 & 0.0 & 0.6 & 2.5 & 0.0 & 1.9 & 3.5 & 0.0 & 2.0 & 1.1 & 0.0 & 0.9 & 2.8 & 0.0 & 1.5 & 1.8 & 0.0 & 1.1 \\
Qwen2-VL-7B           & 2.6 & 0.0 & 1.3 & 1.5 & 0.0 & 0.9 & 0.5 & 0.0 & 0.4 & 6.3 & 0.0 & 3.5 & 3.4 & 1.9 & 3.0 & 0.9 & 0.0 & 0.5 & 2.5 & 0.2 & 1.6 \\
OS-Atlas-4B           & 7.1 & 0.0 & 3.7 & 3.0 & 1.4 & 2.3 & 2.0 & 0.0 & 1.5 & 9.0 & 5.5 & 7.5 & 5.1 & 3.8 & 4.8 & 5.6 & 0.0 & 3.1 & 5.0 & 1.7 & 3.7 \\
ShowUI-2B             & 16.9 & 1.4 & 9.4 & 9.1 & 0.0 & 5.3 & 2.5 & 0.0 & 1.9 & 13.2 & 7.3 & 10.6 & 15.3 & 7.5 & 13.5 & 10.3 & 2.2 & 6.6 & 10.8 & 2.6 & 7.7 \\
CogAgent-18B          & 14.9 & 0.7 & 8.0 & 9.6 & 0.0 & 5.6 & 7.1 & 3.1 & 6.1 & 22.2 & 1.8 & 13.4 & 13.0 & 0.0 & 10.0 & 5.6 & 0.0 & 3.1 & 12.0 & 0.8 & 7.7 \\
Aria-UI               & 16.2 & 0.0 & 8.4 & 23.7 & 2.1 & 14.7 & 7.6 & 1.6 & 6.1 & 27.1 & 6.4 & 18.1 & 20.3 & 1.9 & 16.1 & 4.7 & 0.0 & 2.6 & 17.1 & 2.0 & 11.3 \\
UGround-7B            & 26.6 & 2.1 & 14.7 & 27.3 & 2.8 & 17.0 & 14.2 & 1.6 & 11.1 & 31.9 & 2.7 & 19.3 & 31.6 & 11.3 & 27.0 & 17.8 & 0.0 & 9.7 & 25.0 & 2.8 & 16.5 \\
Claude Computer Use~\cite{bai2025qwen25vl}   & 22.0 & 3.9 & 12.6 & 25.9 & 3.4 & 16.8 & 14.5 & 3.7 & 11.9 & 33.9 & 15.8 & 25.8 & 30.1 & 16.3 & 26.9 & 11.0 & 4.5 & 8.1 & 23.4 & 7.1 & 17.1 \\
OS-Atlas-7B           & \underline{33.1} & 1.4 & 17.7 & 28.8 & 2.8 & 17.9 & 12.2 & 4.7 & 10.3 & 37.5 & 7.3 & 24.4 & 33.9 & 5.7 & 27.4 & \textbf{27.1} & 4.5 & 16.8 & 28.1 & 4.0 & 18.9 \\
UGround-V1-2B         & --  & --  & \textbf{27.4} & --  & --  & 26.7 & --  & --  & \underline{14.6} & --  & --  & 34.3 & --  & --  & 38.3 & --  & --  & \underline{17.9} & --  & --  & 26.6 \\
Qwen2.5-VL-7B         & --  & --  & 26.1 & --  & --  & 24.0 & --  & --  & 13.0 & --  & --  & 31.1 & --  & --  & 45.2 & --  & --  & \textbf{23.5} & --  & --  & 26.8 \\
UI-TARS-2B            & \textbf{47.4} & \underline{4.1} & \underline{26.4} & \textbf{42.9} & \underline{6.3} & \textbf{27.6} & \underline{17.8} & \underline{4.7} & \underline{14.6} & \textbf{56.9} & \underline{17.3} & \textbf{39.8} & \underline{50.3} & \underline{17.0} & \underline{42.6} & \underline{21.5} & \underline{5.6} & 14.3 & \textbf{39.6} & \underline{8.4} & \underline{27.7} \\
\midrule
\textbf{ZonUI-3B(Ours)}          & 24.6 & \textbf{6.2} & 15.7 & \underline{40.9} & \textbf{7.6} & \underline{26.9} & \textbf{31.9} & \textbf{15.6} & \textbf{27.9} & \underline{54.8} & \textbf{18.1} & \underline{38.9} & \textbf{57.0} & \textbf{26.4} & \textbf{50.0} & 19.6 & \textbf{7.8} & 14.2 & \underline{39.2} & \textbf{11.7} & \textbf{28.7} \\
\bottomrule
\end{tabular}}
\end{table*}

\section{Experiments}
\subsection{Experimental Setup and Evaluation}

Our experiments are conducted using a single NVIDIA RTX 4090 GPU (24,GB VRAM). Model training follows a resource-efficient setup based on DeepSpeed ZeRO-2 optimization and FlashAttention (SDPA) to reduce memory overhead and improve attention throughput. Fine-tuning is performed using LoRA with rank 8 and $\alpha = 16$, applied to selected transformer layers \cite{hu2021lora}. Mixed-precision (FP16) training is used throughout, with a batch size of 1 and gradient accumulation steps set to 48. Each training epoch consists of 122 steps.

Training is performed in two stages:

\begin{itemize}
\item \textbf{Stage 1: Cross-Platform Fine-Tuning.}
The model is trained on a mixed dataset comprising web, mobile, and desktop GUI screenshots \cite{lin2024showui, chai2025amex}. A learning rate of 2e-4 is used, and balanced sampling is applied to mitigate platform imbalance, ensuring sufficient exposure to under-represented desktop data.

\item \textbf{Stage 2: High-Resolution Specialization.}
The Stage~1 model is further adapted on a high-resolution subset drawn from UGround’s web-hybrid data \cite{gou2024uground}. This subset introduces diverse screen resolutions and aspect ratios, including mobile-formatted web layouts and full-screen desktop renderings. A reduced learning rate of 5e-5 is used in this stage to stabilize fine-tuning on dense visual inputs.
\end{itemize}

Evaluation is conducted on three GUI grounding benchmarks: ScreenSpot \cite{cheng2024seeclick}, ScreenSpot-v2 \cite{wu2024osatlas}, and ScreenSpot-Pro \cite{li2025screenspotpro}. Each benchmark comprises 1,272 natural language grounding tasks, paired with GUI screenshots sampled from mobile, desktop, and web environments. Success is defined by whether the predicted coordinates fall within the annotated bounding box.

ScreenSpot-v2 improves upon the original benchmark by correcting annotation errors and clarifying ambiguous instructions \cite{wu2024osatlas}, providing a more reliable testbed. ScreenSpot-Pro~\cite{li2025screenspotpro} targets professional, high-resolution interfaces with full-screen layouts from tools such as Photoshop, AutoCAD, and MATLAB. It features smaller UI targets and more visually complex scenes, serving as a realistic evaluation for models operating in demanding desktop settings.

\subsection{Main Results: Accuracy and Model Comparison}

We evaluate ZonUI-3B on three standard GUI grounding benchmarks: ScreenSpot \cite{cheng2024seeclick}, ScreenSpot-v2 \cite{wu2024osatlas}, and the newly proposed ScreenSpot-Pro \cite{li2025screenspotpro}. Despite its compact 3B parameter scale, the model achieves strong performance across all settings.

On \textbf{ScreenSpot}, ZonUI-3B achieves an accuracy of \textbf{84.9\%}, and on the cleaned \textbf{ScreenSpot-v2}, it reaches \textbf{86.4\%}, setting a new benchmark among all sub-4B models. Compared to the previous best 2B-scale model, UI-TARS-2B \cite{qin2025uitars}, ZonUI-3B yields a +2.6\% gain on ScreenSpot and +1.7\% on ScreenSpot-v2. These results indicate that our training strategy and dataset design effectively close the performance gap between small and mid-sized models. Beyond sub-4B models, ZonUI-3B performs on par with several 7B-scale systems, such as Aguvis-7B \cite{xu2024aguvis} and OS-Atlas-7B \cite{wu2024osatlas}. These results suggest that, with proper adaptation and data curation, a 3B model can achieve competitive performance typically associated with models 2$\times$ its size. See Table~\ref{tab:screenspot} for detailed results.

On \textbf{ScreenSpot-Pro}, a challenging benchmark that includes professional software environments (e.g., Photoshop, AutoCAD) with dense visual layouts and smaller UI targets, ZonUI-3B achieves a success rate of \textbf{28.7\%}. This surpasses the strongest prior 2B baseline (27.7\%) and exceeds several 7B models, further validating its robustness in high-resolution, real-world applications. See Table~\ref{tab:screenspot_pro} for details.

These results establish ZonUI-3B as a strong candidate for GUI grounding under constrained resources. The model demonstrates that with a compact architecture and carefully designed training strategy, it is possible to attain competitive accuracy without relying on extensive model scale or massive datasets.

\subsection{Ablation Study}

We conduct a series of ablation experiments to evaluate the effectiveness of individual components in our training pipeline, with a particular focus on sampling strategies, dataset composition, and the proposed two-stage fine-tuning schedule.

To address the inherent imbalance in GUI data across platforms, we compare uniform sampling (1:1:1 across mobile, web, and desktop) with joint training that reflects the original data distribution. As shown in Table~\ref{tab:ablation-sampling}, balanced sampling yields a modest improvement in overall accuracy (+0.9\%) and a slight gain on desktop-specific performance (+1.3\%). While the improvement is not large, it suggests that balancing platform representation during training can help stabilize performance on under-represented high-resolution interfaces \cite{lin2024showui, chai2025amex}.

\begin{table}[ht]
\centering
\caption{Effect of balanced sampling during Stage~1 fine-tuning.}
\label{tab:ablation-sampling}
\resizebox{\columnwidth}{!}{
\begin{tabular}{lcc}
\toprule
\textbf{Sampling Method} & \textbf{SS-Acc. (\%)} & \textbf{SS-Desktop (\%)} \\
\midrule
Joint-Training (no balancing) & 81.9 & 79.4 \\
Balanced Sampling (1:1:1 ratio) & \textbf{82.8} & \textbf{80.7} \\
\bottomrule
\end{tabular}
}
\end{table}

\textbf{Dataset Diversity vs. Volume.}
We evaluate the trade-off between dataset size and diversity by comparing subsets drawn from ShowUI-Web, UGround, and their combinations. Table~\ref{tab:ablation-mix} shows that simply scaling up the number of samples (e.g., 16K vs. 119K) leads to negligible gains (+0.1\%) in overall accuracy. In contrast, merging ShowUI with UGround (24.1K examples total) improves performance beyond either dataset alone, demonstrating the benefits of incorporating multi-resolution and stylistic diversity \cite{gou2024uground}. This finding aligns with our observation that existing GUI datasets contain considerable redundancy, and that well-curated, compact subsets can offer equivalent or superior results \cite{lin2024showui}.

\begin{table}[ht]
\centering
\caption{Comparison of data volume and source diversity.}
\label{tab:ablation-mix}
\renewcommand{\arraystretch}{1.2}
\resizebox{\columnwidth}{!}{
\begin{tabular}{lccc}
\toprule
\textbf{Method} & \textbf{Data Size} & \textbf{SS Acc. (\%)} & \textbf{SS-Desktop / SS-Web (\%)} \\
\midrule
Only ShowUI-Web (small set)         & 16.1K  & 82.8  & 80.7 / 80.0 \\
Only ShowUI-Web (large set)         & 119.4K & 82.9  & 79.5 / 81.1 \\
Only UGround-Web                    & 16.1K  & 82.7  & 81.3 / 80.6 \\
UGround + ShowUI-Web                & 24.1K  & 83.5  & 81.0 / 80.9 \\
\textbf{2-Stage Training (Ours)}    & 24.1K  & \textbf{84.9} & \textbf{84.0 / 81.8} \\
\bottomrule
\end{tabular}
}
\end{table}

\textbf{Effect of Two-Stage Fine-Tuning.}
We further isolate the contribution of our two-stage training strategy. As shown in Table~\ref{tab:ablation-stage}, adding multi-resolution data from UGround to a small ShowUI subset yields minor improvements. However, applying two-stage fine-tuning on the same dataset results in a +3.3\% gain on desktop and +1.8\% on web interfaces—substantially outperforming single-stage alternatives. This confirms that separating general GUI pretraining from resolution-specialized adaptation is a key factor in improving grounding accuracy, particularly on visually complex layouts \cite{gou2024uground, bai2025qwen25vl}.

These results validate the effectiveness of our training methodology. Rather than relying on large-scale data or complex architecture modifications, we show that performance can be significantly improved through targeted sampling, dataset composition, and phased training strategies.

\begin{table}[ht]
\centering
\caption{Impact of UGround data and two-stage fine-tuning.}
\label{tab:ablation-stage}
\renewcommand{\arraystretch}{1.2}
\resizebox{\columnwidth}{!}{
\begin{tabular}{lccc}
\toprule
\textbf{Method} & \textbf{Data Size} & \textbf{SS-Desktop} & \textbf{SS-Web} \\
\midrule
ShowUI-Web only             & 16.1K  & 80.7\% (–)           & 80.0\% (–)           \\
+UGround (multi-resolution) & 24.1K  & 81.0\% (+0.3\%)      & 80.9\% (+0.9\%)      \\
\textbf{2-Stage Training (Ours)}  & 24.1K  & \textbf{84.0\%} (+3.3\%) & \textbf{81.8\%} (+1.8\%) \\
\bottomrule
\end{tabular}
}
\end{table}

\section{Conclusion}

This work introduces ZonUI-3B, a 3B-parameter vision-language model designed for GUI grounding under realistic resource constraints. Despite its compact size, the model delivers competitive accuracy across cross-platform benchmarks, including mobile, web, and desktop scenarios. These results demonstrate that lightweight models, when paired with carefully designed training strategies, can close the performance gap to significantly larger systems \cite{qin2025uitars, xu2024aguvis, wu2024osatlas}.

Our approach emphasizes data and training efficiency over scale. By integrating a compact yet diverse dataset (24K examples) and applying a two-stage fine-tuning strategy \cite{gou2024uground, lin2024showui}, ZonUI-3B achieves state-of-the-art performance among sub-4B models and rivals several 7B-scale baselines. Importantly, this is accomplished using a single consumer-grade GPU, without introducing architectural modifications or external modules \cite{hu2021lora}.

Ablation studies further highlight the importance of data diversity, platform balancing, and resolution-aware adaptation. We observe that increasing data volume alone yields diminishing returns, while a small, well-curated dataset can maintain or exceed the effectiveness of much larger corpora \cite{lin2024showui, chai2025amex}. The proposed training framework enables robust grounding even in high-resolution and visually dense desktop environments—traditionally a weakness for lightweight models \cite{gou2024uground}.

ZonUI-3B provides a practical and reproducible strategy for building deployable GUI agents. This study affirms that model compactness need not compromise grounding performance, and that targeted design choices in data construction and training schedules play a critical role in maximizing capability within constrained computational budgets.

{\small
\bibliographystyle{ieee_fullname}
\bibliography{egbib}

\begin{thebibliography}{10}\itemsep=-1pt

\bibitem{bai2025qwen25vl}
Shuai Bai, Keqin Chen, Xuejing Liu, Jialin Wang, Wenbin Ge, Sibo Song, Kai Dang, Peng Wang, Shijie Wang, Jun Tang, Humen Zhong, Yuanzhi Zhu, Mingkun Yang, Zhaohai Li, Jianqiang Wan, Pengfei Wang, Wei Ding, Zheren Fu, Yiheng Xu, Jiabo Ye, Xi Zhang, Tianbao Xie, Zesen Cheng, Hang Zhang, Zhibo Yang, Haiyang Xu, Junyang Lin, et~al.
\newblock Qwen2.5-vl technical report.
\newblock {\em arXiv preprint arXiv:2502.13923}, 2025.

\bibitem{chai2025amex}
Yuxiang Chai, Siyuan Huang, Yazhe Niu, Han Xiao, Liang Liu, Dingyu Zhang, Shuai Ren, and Hongsheng Li.
\newblock Amex: Android multi-annotation expo dataset for mobile gui agents.
\newblock {\em arXiv preprint arXiv:2407.17490}, 2024.

\bibitem{cheng2024seeclick}
Kanzhi Cheng, Qiushi Sun, Yougang Chu, Fangzhi Xu, Yantao Li, Jianbing Zhang, and Zhiyong Wu.
\newblock Seeclick: Harnessing gui grounding for advanced visual gui agents.
\newblock {\em arXiv preprint arXiv:2401.10935}, 2024.
\newblock Version 2, Feb 23, 2024.

\bibitem{gou2024uground}
Boyu Gou, Ruohan Wang, Boyuan Zheng, Yanan Xie, Cheng Chang, Yiheng Shu, Huan Sun, and Yu Su.
\newblock Navigating the digital world as humans do: Universal visual grounding for gui agents.
\newblock {\em arXiv preprint arXiv:2410.05243}, 2024.
\newblock Accepted to ICLR 2025 (Oral).

\bibitem{hong2023cogagent}
Wenyi Hong, Weihan Wang, Qingsong Lv, Jiazheng Xu, Wenmeng Yu, Junhui Ji, Yan Wang, Zihan Wang, Yuxuan Zhang, Juanzi Li, Bin Xu, Yuxiao Dong, Ming Ding, and Jie Tang.
\newblock Cogagent: A visual language model for gui agents.
\newblock {\em arXiv preprint arXiv:2312.08914}, 2023.
\newblock CVPR 2024 (Highlight).

\bibitem{hu2021lora}
Edward~J. Hu, Yelong Shen, Phillip Wallis, Zeyuan Allen-Zhu, Yuanzhi Li, Shean Wang, Lu Wang, and Weizhu Chen.
\newblock Lora: Low-rank adaptation of large language models.
\newblock {\em arXiv preprint arXiv:2106.09685}, 2021.

\bibitem{li2025screenspotpro}
Kaixin Li, Ziyang Meng, Hongzhan Lin, Ziyang Luo, Yuchen Tian, Jing Ma, Zhiyong Huang, and Tat-Seng Chua.
\newblock Screenspot-pro: Gui grounding for professional high-resolution computer use.
\newblock {\em arXiv preprint arXiv:2504.07981}, 2025.

\bibitem{lin2024showui}
Kevin~Qinghong Lin, Linjie Li, Difei Gao, et~al.
\newblock Showui: One vision-language-action model for gui visual agent.
\newblock {\em arXiv preprint arXiv:2411.17465}, 2024.

\bibitem{qin2025uitars}
Yujia Qin, Yining Ye, Junjie Fang, Haoming Wang, Shihao Liang, Shizuo Tian, Junda Zhang, Jiahao Li, Yunxin Li, Shijue Huang, Wanjun Zhong, Kuanye Li, Jiale Yang, Yu Miao, Woyu Lin, Longxiang Liu, Xu Jiang, Qianli Ma, Jingyu Li, Xiaojun Xiao, Kai Cai, Chuang Li, Yaowei Zheng, Chaolin Jin, Chen Li, Xiao Zhou, Minchao Wang, Haoli Chen, Zhaojian Li, Haihua Yang, Haifeng Liu, Feng Lin, Tao Peng, Xin Liu, and Guang Shi.
\newblock Ui-tars: Pioneering automated gui interaction with native agents.
\newblock {\em arXiv preprint arXiv:2501.12326}, 2025.

\bibitem{wu2024osatlas}
Zhiyong Wu, Zhenyu Wu, Fangzhi Xu, Yian Wang, Qiushi Sun, Chengyou Jia, Kanzhi Cheng, Zichen Ding, Liheng Chen, Paul~Pu Liang, and Yu Qiao.
\newblock Os-atlas: A foundation action model for generalist gui agents.
\newblock {\em arXiv preprint arXiv:2410.23218}, 2024.

\bibitem{xu2024aguvis}
Yiheng Xu, Zekun Wang, Junli Wang, Dunjie Lu, Tianbao Xie, Amrita Saha, Doyen Sahoo, Tao Yu, and Caiming Xiong.
\newblock Aguvis: Unified pure vision agents for autonomous gui interaction.
\newblock {\em arXiv preprint arXiv:2412.04454}, 2024.
\newblock Accepted to ICML 2025.

\bibitem{zhu2025internvl3}
Jinguo Zhu, Weiyun Wang, Zhe Chen, Zhaoyang Liu, Shenglong Ye, Lixin Gu, Hao Tian, Yuchen Duan, Weijie Su, Jie Shao, Zhangwei Gao, Erfei Cui, Xuehui Wang, Yue Cao, Yangzhou Liu, Xingguang Wei, Hongjie Zhang, Haomin Wang, Weiye Xu, Hao Li, Jiahao Wang, Nianchen Deng, Songze Li, Yinan He, Tan Jiang, Jiapeng Luo, Yi Wang, Conghui He, Botian Shi, Xingcheng Zhang, Wenqi Shao, Junjun He, Yingtong Xiong, Wenwen Qu, Peng Sun, Penglong Jiao, Han Lv, Lijun Wu, Kaipeng Zhang, Huipeng Deng, Jiaye Ge, Kai Chen, Limin Wang, Min Dou, Lewei Lu, Xizhou Zhu, Tong Lu, Dahua Lin, Yu Qiao, Jifeng Dai, and Wenhai Wang.
\newblock Internvl3: Exploring advanced training and test-time recipes for open-source multimodal models.
\newblock {\em arXiv preprint arXiv:2504.10479}, 2025.

\end{thebibliography}
}

\end{document}